\documentclass[sigconf, natbib]{acmart}

\AtBeginDocument{%
  }

\usepackage{latexsym}

\usepackage[most]{tcolorbox}
\usepackage{lipsum}

\usepackage{fvextra}
\usepackage{caption}
\usepackage{upquote}

\usepackage[T1]{fontenc}

\usepackage[utf8]{inputenc}

\usepackage{microtype}

\usepackage{amsmath}
\usepackage{xcolor}
\usepackage{graphicx}
\usepackage{multicol}
\usepackage{listings}
\usepackage{appendix}
\usepackage{tablefootnote}
\usepackage{array}
\begin{document}
\title{CF-RAG: A Dataset and Method for Carbon Footprint QA Using Retrieval-Augmented Generation}

\author{Kaiwen Zhao}
\affiliation{%
  \institution{University of Pittsburgh}
  \country{USA}
}

\author{Bharathan Balaji}
\affiliation{%
  \institution{Amazon}
  \country{USA}
}

\author{Stephen Lee}
\affiliation{%
  \institution{University of Pittsburgh}
  \country{USA}
}
\begin{abstract}
Product sustainability reports provide valuable insights into the environmental impacts of a product and are often distributed in PDF format. 
These reports often include a combination of tables and text, which complicates their analysis. The lack of standardization and the variability in reporting formats further exacerbate the difficulty of extracting and interpreting relevant information from large volumes of documents.
In this paper, we tackle the challenge of answering questions related to carbon footprints within sustainability reports available in PDF format. Unlike previous approaches, our focus is on addressing the difficulties posed by the unstructured and inconsistent nature of text extracted from PDF parsing. To facilitate this analysis, we introduce CarbonPDF-QA, an open-source dataset containing question-answer pairs for 1735 product report documents, along with human-annotated answers. Our analysis shows that GPT-4o struggles to answer questions with data inconsistencies. To address this limitation, we propose CarbonPDF, an LLM-based technique specifically designed to answer carbon footprint questions on such datasets.
We develop CarbonPDF by fine-tuning Llama 3 with our training data. Our results show that our technique outperforms current state-of-the-art techniques, including question-answering (QA) systems finetuned on table and text data. 
\end{abstract}

\ccsdesc[500]{Information systems~Language models}



\keywords{Product Carbon Footprint, QA, LLM}


\maketitle

\section{Introduction}

\begin{figure*}[th]
    \centering
    \begin{tabular}{c}
    \includegraphics[width=6in]{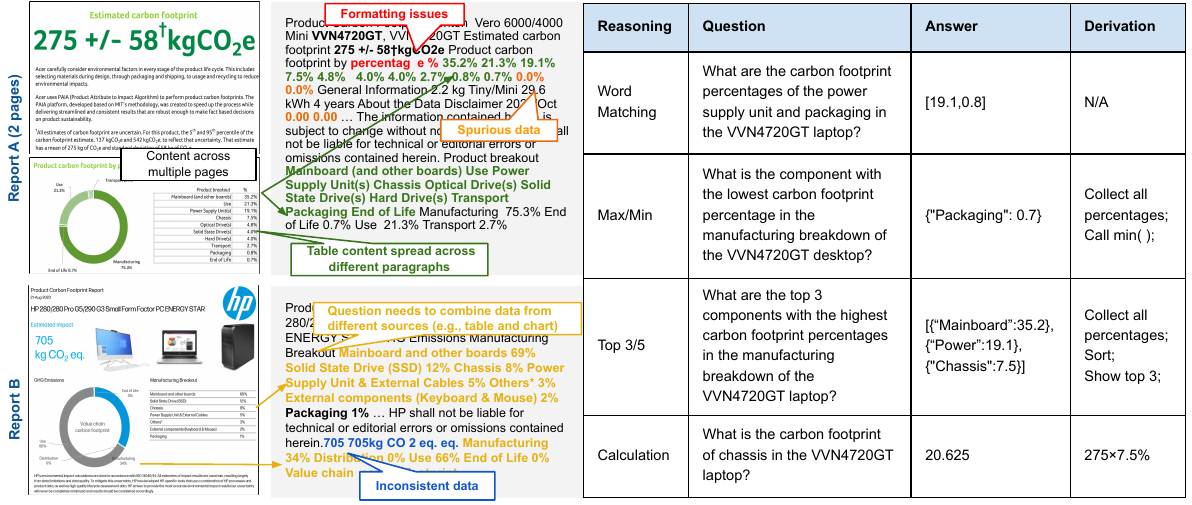} 
    \end{tabular}
    \caption{The CarbonPCF-QA dataset is derived from Product Carbon Footprint (PCF) reports. Examples demonstrate the parsing process used to collect data from the extracted raw text of PDF documents. An accompanying table presents an overview of the various question types included in the dataset.}
    \label{fig:pdf_example}
    
\end{figure*}
Product sustainability reporting provides insights into the environmental impact of products~\cite{sustreport, unreport1}. This reporting is essential not only for regulatory compliance but also for consumers who seek to understand the environmental impact of a product. Furthermore, emission data extracted from these reports enable life cycle assessments and support various models that evaluate environmental impacts and ensure alignment with sustainability commitments~\cite{gupta2022act}. 
To conduct thorough analyses, stakeholders such as regulators and consumers extract data from sustainability reports to perform carbon footprint assessments. These analyses enable the evaluation of a product's environmental impact based on reported data.
However, extracting emissions data from these reports remains largely manual and time-consuming. Moreover, the lack of standardization and the complex format of these reports containing hybrid data --- a mix of tables and text --- present significant challenges for automated extraction and analysis.
Inconsistencies in how data is presented make it difficult to perform numerical reasoning and make it challenging to compare and assess emissions data across different companies or even within the same organization over time.

Recent studies have explored the use of question-answering (QA) techniques for analyzing and extracting numerical data in hybrid formats, such as tables and text, by framing data analysis as questions~\cite{zhu2021tat,chen2022convfinqa}. These approaches use language models to interpret hybrid data and perform numerical reasoning, simplifying the extraction and analysis process~\cite{zhu2024tat}. 
A common approach involves feeding hybrid data and specific questions into language models to generate answers.
However, analyzing hybrid data in carbon sustainability reports presents significant challenges. These difficulties arise because reports are often available as Portable Document Format (PDF) documents, and extracting hybrid data from PDFs is often error-prone. For instance,  although tables may appear structured, PDF does not encode this information as tables, unlike HTML or spreadsheets. Instead, PDFs represent tables as a collection of text and lines placed at specific coordinates without any explicit information about rows or columns. 
This lack of inherent structure makes it difficult to extract and reconstruct tables accurately, as extraction algorithms must infer relationships between text elements based on their positions, which can be complex and unreliable.
As shown in Figure~\ref{fig:pdf_example}, the data extracted from a PDF may appear in a different order than expected, even if it looks sequential in the document.  This happens because the visual layout of the table does not always match a clear, structured format within the PDF file.

The problem is further complicated by variations in how different documents represent tables internally. 
Content may be spread across different pages or sections, making connections between related data loose or unclear. Additionally, hidden text and numbers encoded within the PDF may not be visible but can be read using programs, resulting in spurious or inconsistent data. Existing state-of-the-art QA systems that handle hybrid data generally assume a structured table format, where the content is free from such anomalies~\cite{zhu2024tat}. Thus, these systems may struggle when presented with inconsistent content extracted from PDF documents, where table and text data are represented for visual presentation rather than data analysis. 
Moreover, most QA systems are typically designed to handle reasoning questions over a single table. This limits their effectiveness when dealing with content that spans multiple tables. 

In this paper, we address the challenges associated with the problem of hybrid data extracted from sustainability report PDF documents. Our goal is to answer carbon footprint-related questions based on  data extracted from PDFs, while addressing the challenges posed by inconsistent and loosely connected table and text content. We refer to this data as \textit{inconsistent} because we do not modify the raw text to remove spurious information. Additionally, numbers and text often misalign and can be scattered across multiple paragraphs, complicating direct question answering. To facilitate analysis, we present the CarbonPDF-QA dataset --- an open-domain carbon product report in PDF format. This dataset includes a variety of carbon assessment numerical reasoning questions that require extracting information from text, and tables. We developed this dataset through a combination of automated processes and human verification to ensure its accuracy and reliability. To the best of our knowledge, CarbonPDF-QA is the first dataset specifically designed to include inconsistent data containing unstructured tabular and text content, addressing the unique challenges of analyzing unstructured tabular and text content.



We also develop a specialized, smaller language model, Carbon-PDF, by fine-tuning LLaMA 3, to extract information and answer carbon footprint-related questions directly from sustainability reports in PDF format. Unlike prior work that relies on step-wise reasoning~\cite{zhu2021tat}, our approach generates executable Python programs to compute the final answers. These programs consist of simple, interpretable instructions that perform the necessary calculations to address the questions. We find that generating executable code, instead of just textual responses, enhances accuracy by enabling precise numerical computations. 

Additionally, instead of assuming that the relevant context is readily available for question answering, our approach uses a retrieve-and-generate pipeline.
In this setting, a question is first used to retrieve relevant content from a knowledge base, extracted from the PDFs, and then generate an answer based on that content. This design is useful when relevant information is scattered across large, unstructured data. However, we observed that existing retrievers, such as TF-IDF and Contriever~\cite{izacard2021unsupervised}, often fail to rank the most relevant document at the top, which limits the effectiveness of question answering. To address this, Carbon-PDF incorporates a critic model that re-evaluates and refines the initial retrieval results. The critic assesses the relevance of each candidate document with respect to the question and selects the most contextually appropriate content. This refinement step significantly improves document selection, ensuring that the generator receives high-quality input to produce accurate and well-grounded answers.

In summary, we make the following contributions. 
\begin{itemize}
    \item We introduce CarbonPDF-QA dataset, an open-domain question answering benchmark for carbon product PDF documents that contain unstructured tables and text data.  The dataset comprises 1,735 documents collected from four different companies, with ground truth answers that have been manually verified for accuracy.  These documents reflect real-world reporting challenges, including inconsistent formatting and noisy data, making them valuable for developing robust QA systems. In addition, CarbonPDF-QA includes component-level carbon footprint breakdowns for products, providing fine-grained data for more detailed analysis and question answering.

    \item We develop the CarbonPDF model, a QA system designed to handle the complexities of inconsistent or spurious data extracted from PDF documents. Additionally, our model is built within a retrieve-and-generate (RAG) framework and incorporates a critic module that evaluates and selects the most relevant retrieved content, enabling the generation of accurate and contextually grounded answers.

    \item We conduct extensive experiments and demonstrate that our model outperforms existing state-of-the-art techniques, including GPT and specialized QA systems. Additionally, we perform detailed analyses to showcase the model's capabilities in handling complex numerical reasoning on unstructured table and text data.
\end{itemize}

\section{CarbonPDF-QA Dataset}
\label{Dataset}


\begin{table}[t]
\caption{Statistics of the CarbonPDF-QA dataset}
\centering
\small

\begin{tabular}{l|l!{\vrule width 1.8pt}l|l|l}
\toprule
\multicolumn{2}{c!{\vrule width 1.8pt}}{\textbf{PDF Statistic}} & \multicolumn{3}{c}{\textbf{QA Dataset Summary}} \\\hline
\textbf{Type} & &\textbf{Ques. Type} & \textbf{Train} & \textbf{Test}  \\
\midrule
\# Company & 4 & Word Match & 7105 & 1841 \\\hline
\# File & 1735 & Max/Min & 1863 & 486   \\\hline
 Avg.char./file & 3759 & Top 3/5 & 1242 & 324 \\\hline
Avg.words/file & 539 & Calculation & 4172 & 1093  \\\hline
Avg.pages/file &1.76 & Total Ques. & 14382 & 3744  \\
\bottomrule
\end{tabular}

\label{tab:prompt_dataset}
\end{table}


\begin{table}[t]
\caption{QA Dataset comparison. RC: Reading comprehension. FV: Fact Verification. Arith: Arithmetic.}
    \centering
    \small

    \begin{tabular}{l|l|l|l|l}\toprule
         \textbf{Dataset} & \textbf{\#Doc.}  &\begin{tabular}[c]{@{}l@{}} \textbf{Data}\\\textbf{Source}  \\\end{tabular} &\textbf{Content}&\begin{tabular}[c]{@{}l@{}} \textbf{Ques.}\\\textbf{Type}  \\\end{tabular}\\\midrule
          \begin{tabular}[c]{@{}l@{}}SQuAD \cite{rajpurkar2016squad}
 \\\end{tabular} & 736 & Text & Wikipedia&RC
 \\\hline
 
 \begin{tabular}[c]{@{}l@{}}HybridQA\cite{chen2020hybridqa}
 \\\end{tabular} & \begin{tabular}[c]{@{}l@{}} 13k \\(tables) \\\end{tabular} & \begin{tabular}[c]{@{}l@{}} Text,Table  \\\end{tabular}& Wikipedia &RC
 \\\hline
 
 \begin{tabular}[c]{@{}l@{}}TABFACT\cite{chen2019tabfact}
 \\\end{tabular} & \begin{tabular}[c]{@{}l@{}} 16k \\(tables) \\\end{tabular} & \begin{tabular}[c]{@{}l@{}} Text,Table  \\\end{tabular}  & Wikipedia & FV
 \\\hline
 
          \begin{tabular}[c]{@{}l@{}}TAT-QA\cite{zhu2021tat}\\\end{tabular} & 182 &\begin{tabular}[c]{@{}l@{}} Text,Table \\\end{tabular} & \begin{tabular}[c]{@{}l@{}} Financial \\report \\\end{tabular} & \begin{tabular}[c]{@{}l@{}} RC\\Arith  \\\end{tabular}
          \\\hline
          
          \begin{tabular}[c]{@{}l@{}}\textbf{CaronPDF}\textbf{-QA}
          \\\end{tabular} & \textbf{1735} & \textbf{\begin{tabular}[c]{@{}l@{}} Text,Table,Chart \\\end{tabular}} &\begin{tabular}[c]{@{}l@{}} Carbon \\report \\\end{tabular} & \begin{tabular}[c]{@{}l@{}} RC\\Arith  \\\end{tabular}
          \\\bottomrule 
    \end{tabular}
    \footnotesize
    
    \label{tab:dataset_comparison}
\end{table}

\begin{figure}[t]
\begin{lstlisting}[language=python,caption={Example regular expressions for HP data collection}, label={lst:regex},captionpos=b]
patterns = {
    'ssd': r'Solid State Drive \(SSD\)\s*(\d+(\.\d+)?)%',
    'batteries': r'Batteries\s*(\d+(\.\d+)?)%',
    'chassis': r'Chassis\s*(\d+(\.\d+)?)%',
    'mainboard': r'Mainboard and other boards\s*(\d+(\.\d+)?)%',
    ...
}
\end{lstlisting}
\end{figure}

\begin{figure*}[t]
    \centering
    \includegraphics[width=6.2in]{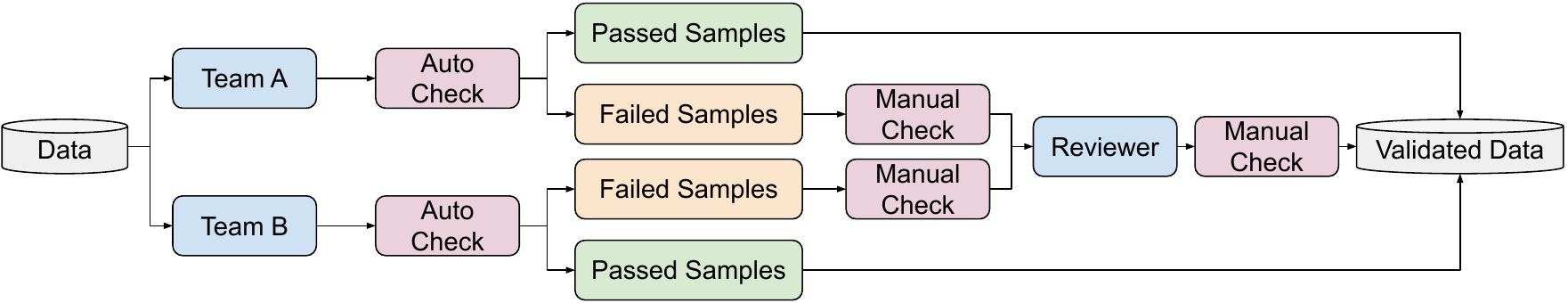}
    \caption{Data validation workflow.}
    \label{fig:validation}
\end{figure*}

\begin{figure}[t]
\begin{lstlisting}[language=python,caption={HP Calculation question program example}, label={lst:program example},captionpos=b]
total_carbon=505.0
manufacturing_percent=0.5
manufacturing_carbon=total_carbon*manufacturing_percent
display_percent=0.24
display_carbon=total_carbon*manufacturing_percent*display_percent
answer=[manufacturing_carbon,display_carbon]
\end{lstlisting}
\end{figure}



\subsection{Data Collection}
Our datasets are derived from computing products' carbon footprint reports, as shown in the left part of Table~\ref{tab:prompt_dataset}. We collected 1,735 PDF reports from the websites of HP \citep{HPLibrary}, Dell \citep{DellLibrary}, Acer \citep{AcerLibrary}, and Lenovo \citep{LenovoLibrary}. Each file contains, on average, around 4k characters and 2 pages. 
In Table \ref{tab:dataset_comparison}, we compare our dataset with existing QA datasets. While most datasets focus on text or table-based data, ours also incorporates text extracted from charts. Additionally, our dataset is sourced from carbon reports and offers a novel contribution to QA research by supporting complex arithmetic reasoning tasks. Unlike prior datasets, which are well-formatted and free from issues, our table data is extracted directly from PDFs without formatting. As a result, rows from the same table may overlap across different paragraphs, adding additional complexity to the reasoning process.

To process these reports, we use the PyMuPDF library \citep{PyMuPDF} to open, parse, and convert the PDF files into text. We developed a Python script to extract product specifications, such as product name, display size, and product weight, as well as carbon-related information, including the total product carbon footprint (PCF) and the carbon footprint percentage of each component in the manufacturing carbon footprint breakdown. 
To accurately identify and extract relevant text and numerical values, we employ regular expressions, which allow us to efficiently locate and retrieve specific data points from the extracted text.
Listing \ref{lst:regex} presents example regular expressions used for parsing carbon reports from HP. In the PCF reports, we observed that the component carbon footprint percentage typically follows the component name or its abbreviation and ends with a \% sign. We also account for decimal points when capturing float values.
If the regular expression is unable to find the expected pattern or detects multiple values, we discard such documents and exclude them from our dataset. Since the document contains only a single instance of each required data point, this approach ensures accuracy and consistency in data extraction.
The extracted text and values are stored in CSV files, which are used to generate the dataset.

\subsection{Dataset Preparation}
\label{sec:data prepare}
\textbf{Question Generation} The dataset includes various question types, shown in the right part of Table~\ref{tab:prompt_dataset}. These range from word-matching questions, where answers can be directly extracted from the PDF file, such as the total product carbon footprint or the carbon footprint percentage of a specific component, which is the "Word Match" question. "Max/Min" question requires identifying the component with either the highest or lowest percentage. "Top 3/5" question involves identifying the top 3 or 5 components that have the largest carbon footprint. The "Calculation" question requires not only evidence extraction from the PDF document that spans different sections but also arithmetic calculations to derive the final answers. The questions in the dataset focus on various aspects of product carbon footprints, including those related to individual components or multiple components of a product. For each product document, we generate, on average, at least 10 questions that can be answered using information from the PDFs.
We wrote a Python script to generate questions based on a predefined question template. A sample template is: "What are the carbon footprints of 
\colorbox[HTML]{fbd9d3}{\texttt{[components]}} in the 
 \colorbox[HTML]{fbd9d3}{\texttt{[product\_name]}}
 \colorbox[HTML]{fbd9d3}{\texttt{[product\_type]}}?" The placeholders are replaced with specific component names (e.g., SSD, display), product names, and product types (e.g., laptop, workstation) from the CSV files, which are extracted from the PDFs.
Each question is paired with a reference document that provides the context for the answer.

\noindent\textbf{Program Generation}
For each question in our dataset, we provide a corresponding Python program that computes the correct answer. These programs are generated using predefined templates tailored to specific question types. The templates follow a consistent structure to ensure interpretability and reliability. 

Listing~\ref{lst:program example} shows an example of a generated program for an HP product, where the question asks for both the manufacturing emissions and the display emissions. The Calculation question program begins by assigning the total Product Carbon Footprint (PCF) value, which serves as the base for all calculations. If the question pertains solely to the total PCF, the program returns that value directly and terminates. If the question involves a breakdown, it proceeds by retrieving the manufacturing percentage, which is always relevant for HP products. If the query focuses only on manufacturing emissions, the program computes the product of the PCF and manufacturing percentage and returns the result. For more specific queries involving individual components, the program retrieves each component’s percentage within the manufacturing footprint and calculates the component-level emissions by multiplying the PCF, manufacturing percentage, and the component’s percentage. All results are returned as a list.

For Max/Min or Top 3/5 questions, the program uses a dictionary of component-percentage pairs. It applies sorting or aggregation functions to identify the most or least significant contributors, returning them as key-value pairs in a list. For Word Match questions, which typically involve identifying multiple absolute footprint values or percentages explicitly mentioned in the reference, the program directly returns a list of those values without further computation. These templates ensure consistent, interpretable, and accurate program outputs across all question types.

To ensure correctness, each program is executed via a validation script, which compares the program’s output to the annotated ground truth in the CSV file. This step confirms both logical and numerical consistency. Our dataset also includes multi-answer questions, such as those asking for the carbon footprint of an HDD and a chassis. In such cases, the final output is a list, with the order of values corresponding exactly to the order of components mentioned in the question to preserve semantic alignment. Finally, we split the dataset into training and test sets using an 80/20 ratio, ensuring strict document-level separation. This prevents any overlap of product carbon reports between the two sets and avoids data leakage during evaluation.

\noindent \textbf{Data Validation}
Figure~\ref{fig:validation} illustrates our data validation process, which was conducted as part of a semester-long Capstone project. Four students voluntarily participated in the project, selecting it from a list of approved Capstone topics based on their interest in data analysis and sustainability. The project focused on the collection, visualization, and analysis of Product Carbon Footprint (PCF) data. Students received academic credit for their contributions and attended weekly meetings for progress updates, technical support, and code reviews. The dataset developed through this collaboration served as a core resource for this research.

The validation process involved dividing the students into two teams, each responsible for running automated verification scripts on the collected data. These scripts checked whether the sum of component percentages fell within 99\%–101\% and whether a product's PCF value deviated by more than \(2 \times \) MAE from the dataset mean. Samples that passed both checks were marked as validated. Of the entire dataset, 24 samples failed the component sum check, and 56 failed the PCF deviation check. Both teams manually reviewed all failed cases. The PCF deviations were found to reflect valid values as reported in the source documents, so these entries were retained. For the 24 sum-check failures, teams carefully examined the documents and corrected errors when discrepancies were found. All corrected entries were subsequently reviewed by the first author and revalidated before being included in the final dataset. This structured validation process ensured high data quality and reliability, combining automated heuristics with human oversight.

\begin{figure*}[h]
    \centering
    \includegraphics[width=6.2in]{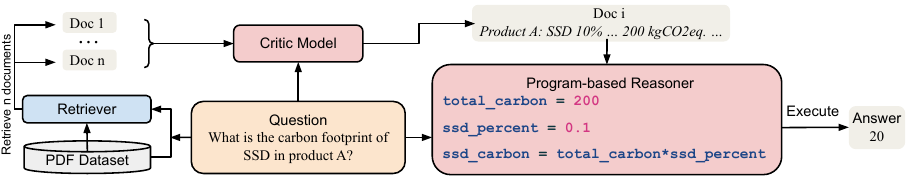}
    \caption{Overview of CarbonPDF.}
    \label{fig:main_design}
\end{figure*}

\section{CarbonPDF Design}
\label{Design}
\subsection{Overview}
A key design goal of CarbonPDF is to provide accurate, fact-based answers to user queries. However, prior work shows that state-of-the-art LLMs often struggle with maintaining factual accuracy~\cite{mallen2022not}. To mitigate this issue, we incorporate Retrieval Augmented Generation (RAG) techniques into our design strategy, leveraging their success in reducing factual errors in knowledge-intensive tasks. Unlike prior work such as TAT-LLM~\cite{zhu2024tat}, which assumes that the retrieved context from the dataset is always correct, our approach acknowledges that the output from retrievers may not always provide the correct context --- a challenge we address in this work.

Figure \ref{fig:main_design} illustrates our approach's key components and overall workflow. For a given question, the retriever first retrieves relevant documents from the PDF database. Next, the critic model finds the most relevant PDF document containing the answer. 
The retrieved reference text, which includes unstructured PDF data, is then combined with the question and a set of instructions to guide the carbon model in its reasoning process to derive the final answer.
The program-based reasoner generates a program that produces the final answer. A program interpreter then executes the necessary calculations to generate the response.  



\begin{figure}[t]
\centering
  \includegraphics[width=2.8in]{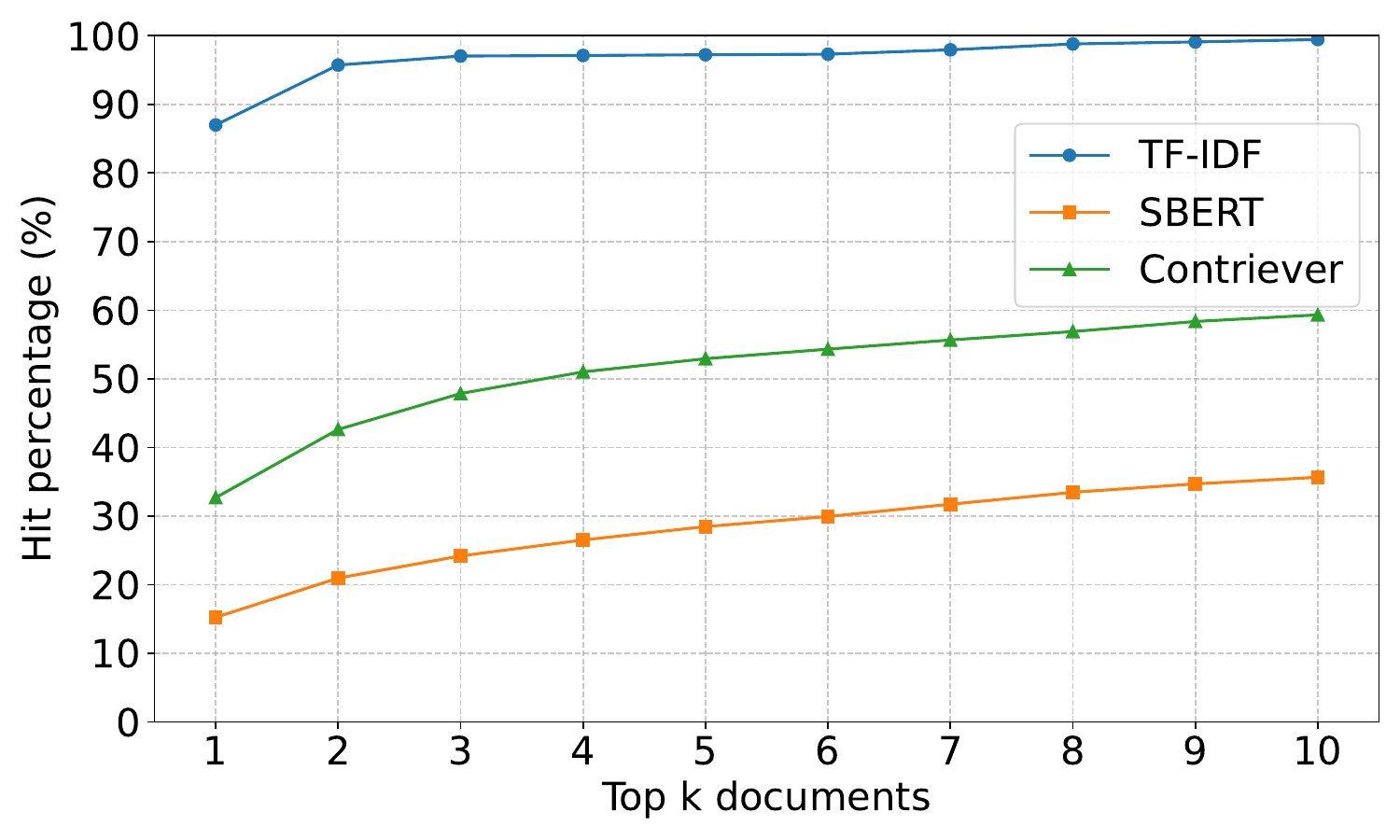}
  \caption{Comparison of retrievers}
  \label{fig:retriever}
\end{figure}


\subsection{Retriever}
Document retrieval has traditionally identified relevant documents through keyword matching. 
Recently, neural network-based approaches, such as Contriever~\cite{izacard2021unsupervised}, which utilize neural embeddings for retrieval, have been introduced and employed in models like Self-RAG~\cite{asai2023self}.
SBERT embedding is another neural embedding commonly used for similar texts matching, which is adapted in CaML \cite{balaji2023caml, reimers2019sentence}.
However, our work uses Term Frequency-Inverse Document Frequency (TF-IDF) embedding as it has a high hit rate for our use case. For each question, we consider it a hit if the correct document appears in the top-k retrieved documents.

Figure \ref{fig:retriever} compares the performance of different retrievers on the test set. 
The figure shows that as k increases, the likelihood of retrieving the correct document within the top-k results also increases.
TF-IDF achieves the highest match percentage, reaching nearly 100\% at k=10, indicating that a relevant document is almost always found within the top 10 results. While we use TF-IDF at present, our approach is flexible enough to support other retrieval techniques.

To retrieve the relevant documents, we first convert the entire document corpus into TF-IDF embeddings using the \textsl{sklearn.TfidfVectorizer} function. This function transforms each document into a vector of numerical values, where each dimension represents a term in the corpus, and the value in each dimension corresponds to the term's TF-IDF score.
When CarbonPDF receives a question, it also converts this question into a TF-IDF vector. We then compute the cosine similarity between the question vector and the TF-IDF vectors of the documents in our corpus. This similarity metric produces a ranking of documents based on their relevance to the question.
The retriever $r(n)$ then selects $n$ documents with the highest similarity scores and passes them to the critic model.

However, even with TF-IDF, the top-1 retrieval hit rate is only around 87\%, meaning that the most relevant document is not always ranked first. To address this limitation, we introduce the critic model that selects the most appropriate document from the top-$k$ candidates, which is discussed in detail in the following section.

\subsection{Critic Model}
The goal of the critic model $\mathcal{C}$ is to identify the document that selects the most relevant context for answering a given question in the top-$n$ retrieved documents from the retriever. Let $\mathcal{D}{critic} = {(q_i, r(q_i, k))}_{i=1}^N$ be our training dataset, consisting of $N$ question samples, where each question $q_i$ is paired with $k$ documents $r(k)$ retrieved based on semantic similarity. In $\mathcal{D}_{critic}$, we label the ground truth document $d$ from the $k$ retrieved documents for each question $q_i$. These ground truth document-question pairs are constructed in the question generation step for each product based on its corresponding product carbon report, which serves as the gold reference document containing the necessary information to answer the question accurately.

We fine-tune the critic model on $\mathcal{D}_{critic}$ using a conditional language modeling objective, maximizing the likelihood:

\begin{equation} \label{eq:critic model}
\max_{\mathcal{C}} \quad \mathbb{E}_{((q,r(q, k)),d) \sim \mathcal{D}_{critic}} \log p_{\mathcal{C}} (d| q, r(q, k))
\end{equation} 

The critic model $\mathcal{C}$ identifies and selects the most appropriate document $d$ for answering question $q$. By training on realistic document-question pairs derived from carbon footprint reports, the model adapts to inconsistencies and noise commonly found in real-world sustainability disclosures.





\subsection{Program-based Reasoner}
LLMs often struggle with reasoning questions that involve complex calculations~\cite{lewkowycz2022solving}. Recently, program-aided language models have demonstrated effectiveness in overcoming these challenges by leveraging programmatic reasoning~\cite{gao2023pal}. This approach improves the model's ability to perform complex calculations and produce accurate answers. Building on this insight, we fine-tune CarbonPDF LLM to generate a Python program to compute the results based on the reference.

The final program generated by CarbonPDF varies based on the type of question and the document's content. Some carbon documents provide the total carbon footprint along with lifecycle breakdowns (e.g., manufacturing, end-of-life, and transport) and detailed breakdowns for individual components (e.g., HDDs, chassis). Other documents may report the carbon footprint of individual components directly without offering a comprehensive lifecycle breakdown.

Similarly, the complexity of the questions may also affect the program generated. For instance, questions requiring detailed calculations for individual components, which depend on factors like the manufacturing process, involve more complex reasoning. To handle such complexity, CarbonPDF employs a multistep approach in its generated programs. Unlike single-step calculations, we store intermediate values in variables and then perform necessary multiplications to derive the answer. In situations with multiple answers, CarbonPDF produces the final answers as a list, maintaining the specified question order.

\begin{figure}[t]
\begin{lstlisting}[caption={Critic model training prompt template}, label={lst:critic prompt},captionpos=b]
You will be provided with a question and several reference texts, each identified by a unique ID. Your goal is to analyze these references and identify which one contains the information needed to answer the question. If a reference text suggests that it provides the necessary information, respond with its corresponding ID. If multiple references apply, respond with a list of their IDs. If none of the references apply, respond with [-1]. Ensure the final output is a list.
### Question: {question}
### Reference 1: {reference[1]}
...
### Reference n: {reference[n]}
### Output: {relevant reference IDs}
\end{lstlisting}
\end{figure}

\begin{figure}[t]
\begin{lstlisting}[caption={Program-based reasoner training prompt template}, label={lst:training prompt},captionpos=b]
You'll be provided with some questions and a reference. Based on the reference, generate the Python program to compute and answer the questions. The program is enclosed by triple backticks. The final answer in the program is of list type.
### Question: {question}
### Reference: {reference text}
### Program:
```
{program}
```
\end{lstlisting}
\end{figure}

\subsection{Training}
We fine-tune our CarbonPDF model based on Llama 3$_{8\mathrm{B}}$\cite{Llama38B}, training both the critic model (Listing~\ref{lst:critic prompt}) and the program-based reasoner (Listing~\ref{lst:training prompt}) separately for 2.5 days each using two NVIDIA RTX 6000 Ada GPUs. The learning rate is set to 2.5e-5, with a per-device training batch size of 8 and gradient accumulation steps of 4. The total number of training epochs is 3. We employ Low-Rank Adaptation (LoRA)\cite{hu2021lora} during training, with 4-bit quantization to reduce memory usage. The paged Adam optimizer\cite{kingma2014adam}, adapted for quantized training, is used to further optimize convergence.

The prompts used in training mirror the inference setup. For the program-based reasoner, the prompt requires the model to generate a Python program enclosed in triple backticks, ensuring structured output and compatibility with program execution. For the critic model, the prompt presents a list of candidate reference texts, each with a unique ID, and asks the model to return only the ID(s) of the relevant reference(s). This design reduces the sequence length, conserves GPU memory, and improves output precision by avoiding noisy or verbose generations.

\section{Evaluation Methodology}



\subsection{Baseline Techniques}
\noindent\textbf{Baselines without LLM.}
ACT~\citep{gupta2022act} and CaML~\citep{balaji2023caml} are model-based carbon estimation techniques that do not rely on large language models (LLMs). ACT calculates the carbon footprint of each component within computer systems using detailed product manufacturing information. On the other hand, CaML associates product names with North American Industry Classification System (NAICS) codes to estimate a carbon footprint per dollar at the industry sector level. Given that CaML consistently provides the same estimate for `Electronic Computer Manufacturing,’ we assume a default price for computing products to calculate the overall carbon footprint. These models are evaluated by estimating and comparing the total carbon footprint of products against ground truth values. Since they do not process natural language input or perform reasoning, they cannot handle question-answering tasks directly and are only applicable to product-level footprint estimation.

\noindent\textbf{Few-shot without RAG.}
We use Gemini-2.0-flash as our few-shot baseline without RAG~\cite{gemini-2.0-flash}. The model is accessed via the Google API and prompted with four manually selected in-context examples covering different question types. It answers questions solely based on its internal knowledge, without access to external documents. We randomly choose four representative samples, each corresponding to a unique combination of company and question type.

\noindent\textbf{Few-shot with RAG.}
We evaluate Self-RAG~\cite{asai2023self}, DeepSeek-R1-Distill-Llama-8B~\cite{guo2025deepseek}, GPT-4o~\cite{hurst2024gpt}, and Gemini-2.0-flash as few-shot baselines with RAG. Self-RAG retrieves relevant documents and guides the LLM to generate the best possible answer. We provide each question along with the top-1 reference text retrieved by the retriever to DeepSeek-R1, GPT-4o, and Gemini-2.0-flash in order to evaluate their performance on the CarbonPDF-QA dataset without fine-tuning. These models are prompted to directly produce the final answer. In contrast, GPT-4o$_{\text{Program}}$ and Gemini-2.0-flash$_{\text{Program}}$ serve as program-based baselines, where the models are asked to generate a Python program that computes the answer, rather than returning the answer directly. This distinction allows us to assess the effectiveness of program-based reasoning compared to direct answer generation in few-shot settings.

\noindent\textbf{Fine-tuned with RAG.}
We fine-tuned Llama3$_{8B}$, TAT-LLM~\cite{zhu2024tat}, and our CarbonPDF on the dataset. TAT-LLM was originally fine-tuned on Llama2$_{7B}$. For a fair comparison, we fine-tuned it on Llama3$_{8B}$. Note that TAT-LLM employs a step-wise pipeline and requires well-formatted tables and text to answer questions. Since our data lacks formatted tables, we used the reference text as a replacement within their prompt template for fine-tuning. Llama3 is fine-tuned to directly generate final answers given the reference context. Llama3${_{Program}}$ is trained to produce a program to compute the answer. Finally, our CarbonPDF${_{Critic+Program}}$ model incorporates both the critic model to select the most relevant document and a program-based reasoner to generate a program solution.

\subsection{Metrics}
\noindent\textbf{Root Mean Squared Error (RMSE)} measures the square root of the average squared differences between predicted and actual values. Because it squares the errors before averaging, RMSE gives more weight to larger errors, making it sensitive to outliers \cite{chai2014root}.

\noindent\textbf{Mean Absolute Error (MAE)} is the average of the absolute differences between predicted and actual values. It treats all errors equally, providing a more balanced view of overall model error without disproportionately emphasizing large deviations \cite{willmott2005advantages}.

\noindent\textbf{Exact Match (EM)} is defined as the percentage of predictions that exactly match the ground truth~\cite{rajpurkar2016squad}. In other words, it checks whether the predicted output is identical to the expected output. For questions with multiple answers, the model is required to match all gold standard answers exactly, including their order, to be considered correct.

\section{Results}
\label{Results}

\begin{table}[t]
\caption{Baseline performance comparison.}
    \centering
    
    \small
    \begin{tabular}{l|l|l|l}\toprule
         \textbf{Techniques} & \textbf{RMSE} & \textbf{MAE} & \textbf{EM}\\\midrule
         \multicolumn{4}{c}{\textbf{Baselines without LLM}} \\\hline
         \begin{tabular}[c]{@{}l@{}}ACT\cite{gupta2022act}\end{tabular} & 486.83 & 323.80 & 0.00 \\\hline 
         \begin{tabular}[c]{@{}l@{}}CaML\citep{balaji2023caml}\end{tabular} & 435.89 & 230.70 & 0.28  \\\hline
         \multicolumn{4}{c}{\textbf{Few-shot without RAG}} \\\hline
         \begin{tabular}[c]{@{}l@{}}Gemini-2.0-flash \cite{gemini-2.0-flash}
         \end{tabular}  &  76.61 & 71.98 & 0.51   \\\hline

         \multicolumn{4}{c}{\textbf{Few-shot with RAG}} \\\hline
\begin{tabular}[c]{@{}l@{}}Self-RAG\cite{asai2023self}\end{tabular}  & 43.55 & 35.44 &  22.70\\\hline
\begin{tabular}[c]{@{}l@{}}DeepSeek-R1\cite{guo2025deepseek}\end{tabular}  &  28.20 & 21.98 & 35.87   \\\hline
GPT-4o\cite{hurst2024gpt} & 13.37 & 10.20 & 49.20  \\\hline
GPT-4o$_{Program}$ & 7.33 & 5.97 & 62.18  \\\hline
Gemini-2.0-flash &  10.37 & 8.12 & 56.52   \\\hline
Gemini-2.0-flash$_{Program}$ &   6.84 &  5.60  & 65.81   \\\hline
         
         \multicolumn{4}{c}{\textbf{Fine-tuned with RAG}} \\\hline
         \begin{tabular}[c]{@{}l@{}}Llama 3\cite{Llama38B}\end{tabular} &  4.01 & 3.43 & 59.70   \\\hline
         \begin{tabular}[c]{@{}l@{}}Llama 3$_{Program}$\end{tabular} &   2.45 & 2.19 & 83.92   \\\hline
        \begin{tabular}[c]{@{}l@{}}TAT-LLM\cite{zhu2024tat}\end{tabular}
         &  2.92 & 2.52 & 64.13  \\\hline
        \textbf{CarbonPDF$_{Critic+Program}$}&  \textbf{0.78} & \textbf{0.69} & \textbf{93.70}
         \\\bottomrule
    \end{tabular}
     \footnotesize
     
    \label{tab:baseline_comparison}
\end{table}

\subsection{Baseline Performance}
Table~\ref{tab:baseline_comparison} compares CarbonPDF with a range of baseline techniques. Our technique consistently outperforms all baselines, achieving the lowest RMSE (0.78) and MAE (0.69), and the highest EM (93.70). Model-based approaches such as ACT and CaML show high RMSE and MAE due to their reliance on general carbon estimates and default values, which lack customization for specific questions. Consequently, their Exact Match (EM) scores are close to zero. LLM baselines without RAG---such as Gemini-2.0-flash---achieve lower RMSE and MAE because they can draw from broad pretraining knowledge. However, they still exhibit significant errors in exact matching, with EM values below 1\%. This underscores the need for data augmentation to improve model accuracy and robustness.

RAG-based approaches demonstrate improved performance by incorporating retrieved textual evidence, enabling more context-aware answers. Gemini-2.0-flash$_{\text{Program}}$ performs the best among few-shot models with 65.81\% EM, benefiting from both document grounding and structured reasoning via code generation. Prompting the LLM to generate a program for solving the question improves answer accuracy, as it shifts the computational burden to a program executor, which is significantly more precise in arithmetic tasks. We also evaluate Self-RAG, which retrieves and uses relevant text spans for answering. While it performs better than non-RAG models, it struggles with discrete numerical reasoning, resulting in a relatively low EM of 22.70\%.

Fine-tuned models generally outperform few-shot approaches due to task-specific optimization. TAT-LLM achieves strong performance, but still falls short in handling inconsistencies and formatting irregularities common in PDF-based reports. It relies on clean tabular-text alignment, which is often absent in realistic datasets. Our model achieves the best performance by integrating a program reasoner with a critic model. The critic helps validate program outputs, enhancing overall robustness. Compared to the best few-shot RAG baseline with program generation (Gemini-2.0-flash$_{\text{Program}}$), our model reduces RMSE and MAE by factors of 8.8 and 8.1, respectively, and improves EM by 27.9\%. This substantial margin highlights the difficulty of reasoning over real-world sustainability documents and the effectiveness of our full pipeline in addressing those challenges. Unlike prior models that need clean and formatted input, CarbonPDF can handle the inconsistent formats seen in actual PDF reports, making it more suitable for practical applications in sustainability and compliance analysis.

\begin{table}[t]
\caption{Performance on different question types.}
    \centering
 \small
    
    \begin{tabular}{l|c|c|c|c}\toprule
         \textbf{Type} & \textbf{\# Question} & \textbf{RMSE} & \textbf{MAE} & \textbf{EM} \\\midrule
         Word Match &  1841 & 1.02 & 0.97 & 94.41  \\\hline
         Max/Min &   486 & 0.54 & 0.46 & 95.27 \\\hline
         Top 3/5 &  324 & 0.40 & 0.30 & 92.59  \\\hline
         Calculation &  1093 & 0.61 & 0.43 & 92.13 
         \\\bottomrule
    \end{tabular}
    \footnotesize
    
    \label{tab:question-type}
\end{table}
\subsection{Performance on Different Question Types}
Table~\ref{tab:question-type} summarizes our CarbonPDF performance across different question types shown in Figure~\ref{fig:pdf_example}. The model performs well across all categories, consistently achieving over 92\% Exact Match (EM). Max/Min questions yield the best results with 0.54 RMSE, 0.46 MAE, and 95.27\% EM, due to their simpler reasoning and fewer question variants, which reduce the potential for errors. Word-matching questions exhibit the highest RMSE (1.02) and MAE (0.97), as they span a broader range of value types, such as absolute emissions and percentages, introducing more variability. Their high frequency (over 1,800 questions) also makes them more susceptible to subtle numerical inaccuracies.

Calculation questions involve multi-step arithmetic reasoning. While the model achieves strong performance with 0.61 RMSE and 0.43 MAE, the EM drops to 92.13\%, the lowest among all types. This reflects the compounded difficulty of arithmetic operations in noisy, unstructured documents. The Top-3/5 ranking questions also maintain robust performance, showing the model's ability to interpret component footprint rankings based on numerical values. Overall, CarbonPDF demonstrates strong generalization across diverse question types despite variability in format and reasoning complexity.



\begin{table} [t]
\caption{Performance on multi-answer questions.}
    \centering
\small
    
    \begin{tabular}{l|c|c|c|c}\toprule
         \textbf{\#Answer} & \textbf{\#Question} & \textbf{RMSE} & \textbf{MAE} & \textbf{EM}\\\midrule
         1 &  1148 & 1.57 & 1.57 & 95.91 \\\hline
         2 &  878 & 0.37 & 0.27 & 94.53 \\\hline
         3 &  857 & 0.43 & 0.28 & 93.00 \\\hline
         4 &  699 & 0.51 & 0.35 & 90.99 \\\hline
         5 &  162 & 0.46 & 0.32 & 88.89  
         \\\bottomrule
    \end{tabular}
    \footnotesize
    
    \label{tab:number_q}
\end{table}

\subsection{Performance on Multi-answer Questions}
We now analyze the impact of questions requiring multiple answers. Table~\ref{tab:number_q} shows the results as we vary the number of answers per question. As the number of required answers increases, the EM score gradually decreases due to the added complexity in evidence extraction and carbon modeling. The chance of error increases when the model is required to return multiple exact answers, as opposed to a single answer, due to compounded prediction risk. Ensuring the correctness of all returned values, including their order and formatting, becomes more difficult with more targets.

Most multi-answer questions involve a mix of calculation and word-matching types, which can introduce ambiguity or minor deviations that lead to EM mismatches. Interestingly, while single-answer questions achieve the highest EM (95.91\%), they also exhibit the highest RMSE (1.57) and MAE (1.57), primarily because they include total carbon footprint queries with large numerical ranges, amplifying absolute error. Although questions requiring five answers have the lowest EM (88.89\%), their overall RMSE and MAE remain low, suggesting that even when the model misses one or two values, its numerical predictions remain close to the ground truth. This demonstrates CarbonPDF’s resilience in complex multi-answer scenarios.


\begin{table}[t]
\caption{Ablation analysis of CarbonPDF.}
    \centering
    \small
    
    \begin{tabular}{l|l|l|l}\toprule
         \textbf{Task} & \textbf{RMSE} & \textbf{MAE} & \textbf{EM}\\\midrule
          \begin{tabular}[c]{@{}l@{}}Retriever+Few-shot
\\\end{tabular} &  833.09 & 668.10 & 23.77 \\\hline
          \begin{tabular}[c]{@{}l@{}}Retriever+Fine-tuned\\\end{tabular} &   2.45 & 2.19 & 83.92 \\\hline
          \begin{tabular}[c]{@{}l@{}}Retriever+Critic\\+Fine-tuned w/o\\ Program Reasoning
          \\\end{tabular} &  2.39 & 1.96 & 66.83
          \\\hline
          \begin{tabular}[c]{@{}l@{}}\textbf{Retriever+Critic}\\\textbf{+Fine-tuned}
          \\\textbf{(CarbonPDF)}
          \\\end{tabular} & \textbf{0.78} & \textbf{0.69} & \textbf{93.70}
          \\\bottomrule
    \end{tabular}
    \footnotesize
    
    \label{tab:task_bk}
    \vspace{-0.1in}
\end{table}

\subsection{Ablation Study}
We conduct an ablation study to evaluate the impact of various components. First, we analyze the effectiveness of few-shot learning in our pipeline. Few-shot learning involves providing a small number of examples at inference time to guide the desired completion, which has been shown to perform well in some tasks~\cite{brown2020language,gautier2022few}. Thus, we replace the fine-tuning step in CarbonPDF with a few-shot approach (Llama 3), where we provide four examples to derive the program for a given question and reference. Table~\ref{tab:task_bk} shows the results of this approach. We observe that the few-shot technique does not perform well with large RMSE (833.09) and MAE (668.10). This is consistent with prior work that indicates few-shot methods struggle with complex reasoning tasks~\cite{brown2020language,asai2023self}. Additionally, we found that few-shot prompting fails to generalize well across varied reporting formats, limiting its robustness.

The results in the second row are evaluated without the critic model. Comparing them with the last row (CarbonPDF), which utilizes the critic model, we observe that CarbonPDF achieves approximately 10\% higher EM while also reducing RMSE and MAE by a factor of around 3. This shows that the critic model enhances overall accuracy by providing more relevant documents to CarbonPDF. In addition, we evaluated the LLM fine-tuned without the program-based reasoning step. In this variation, we trained it to generate the final answer directly without using the program. CarbonPDF with program-based reasoning improves EM by approximately 27\% compared to the version without it. This underscores the importance of program-based reasoning, as it shifts the computational burden to the program interpreter, leading to significantly more accurate arithmetic results.

\section{Related Work}
\noindent{\bf QA Datasets}
There are numerous existing QA datasets designed to evaluate reasoning over structured and unstructured data. For structured data, such as knowledge bases (KBs) and tables \cite{talmor2018web, chen2019tabfact}, notable examples include Complex Web Questions, which extends WebQuestions by introducing multi-hop compositional questions that require reasoning over a KB and retrieving evidence from the web. TabFact is a large-scale dataset for table-based fact verification, where the task is to verify natural language statements against tabular data.
For text-based QA \cite{rajpurkar2016squad, dunn2017searchqa, dua2019drop}, SQuAD remains a foundational benchmark that focuses on extractive QA over Wikipedia passages, requiring models to locate span-based answers in context. 
SearchQA introduces noise and retrieval challenges by building questions around real Jeopardy! clues and snippets retrieved via Google, simulating a realistic QA scenario. 

In the multi-hop QA domain \cite{yang2018hotpotqa, chen2020hybridqa}, HotpotQA requires models to combine evidence from multiple supporting documents and also provide supporting facts, making it an interpretable benchmark. 
HybridQA further expands the reasoning space by requiring integration of tabular and textual content, encouraging models to navigate between heterogeneous data formats.
Hybrid datasets\cite{zhu2021tat, zhu2024tat} such as TAT-QA integrate financial reports that include both textual and tabular information, focusing on complex reasoning tasks. TAT-LLM refines this idea by providing high-quality annotations to train large language models on structured-discrete reasoning.
Our CarbonPDF-QA dataset stands apart by including inconsistent or spurious data extracted from PDFs, reflecting the challenges of real-world document processing.
Furthermore, the tables in our dataset are not well-structured, with column values that may span different paragraphs, complicating data analysis.

\noindent{\bf QA Reasoning}
Numerous studies have explored question answering (QA) systems, including those leveraging Retrieval-Augmented Generation (RAG) to enhance large language models (LLMs) with external context~\cite{izacard2021unsupervised, wei2022chain, gao2023pal, guu2020retrieval, lewis2020retrieval, asai2023self}. These approaches have significantly improved the factual accuracy and contextual relevance of LLM outputs. However, LLMs continue to face challenges when it comes to complex reasoning tasks, particularly in the domain of numerical reasoning. To address this, recent research has focused on numerical reasoning across various data modalities, including tables, text, and charts~\cite{zhu2021tat, kim2021donut, li2022learning, zhu2022towards, zhou2022unirpg, li2023dyrren, wei2023multi, han2023chartllama}. UNIRPG proposes a unified reading–program generation framework that translates text and tables into executable programs, improving compositional and numerical reasoning in QA. ChartLlama focuses on visual QA over chart images, emphasizing multi-modal numerical reasoning that combines vision and language understanding. These studies demonstrate the importance of aligning structured and semi-structured information for accurate inference. Notably, financial QA datasets such as FinQA~\cite{chen2021finqa} and FinLLMs~\cite{yuan2024finllms} illustrate the challenges and opportunities of reasoning over domain-specific documents with embedded numerical content.

Despite these advancements, the application of numerical reasoning techniques to real-world, unstructured, and often inconsistent data, such as that extracted from PDF documents, remains relatively underexplored. Unlike curated benchmark datasets, real-world PDFs frequently contain fragmented, noisy, or irregularly formatted data, posing significant hurdles for automated reasoning. Our work addresses this gap by focusing on the challenges posed by sustainability reports, which are typically published as PDFs.

\noindent{\bf Carbon Footprint Analysis}
Companies frequently employ Lifecycle Assessment (LCA) methodologies to evaluate the environmental impact of their products throughout the entire lifecycle, from raw material extraction to end-of-life disposal~\cite{hauschild2018life}. Tools such as GaBi and SimaPro are widely used to conduct these assessments, generating detailed environmental analyses that are often incorporated into corporate sustainability reports~\cite{silva2017important}. However, LCA methods typically require substantial manual effort and rely on comprehensive, high-quality input data—much of which is proprietary and not publicly disclosed.

In response to these limitations, recent research has explored automating carbon footprint analysis using data-driven approaches~\cite{gupta2022act, balaji2023caml}. These methods leverage publicly available information, including industry averages and inferred estimates, to produce scalable assessments. While more accessible, these approximations often sacrifice accuracy compared to traditional LCA, which benefits from granular, product-specific data.

Despite progress in sustainability analytics, the integration of question-answering (QA) systems into this domain remains largely unexplored. In particular, the use of QA models to extract and reason over carbon footprint information within unstructured sustainability reports has received little attention. To the best of our knowledge, our work is the first to apply QA systems for carbon footprint assessment using real-world sustainability disclosures.

\section{Discussion}

Our specialized CarbonPDF model, despite its smaller size, outperforms state-of-the-art few-shot models such as GPT-4o and Gemini-2.0-flash on our domain-specific CarbonPDF-QA task. 
This performance is achieved with substantially lower training cost and resource consumption. The results show that smaller, fine-tuned models can effectively extract data from unstructured reports without the computational overhead associated with large language models.

One key limitation of current PDF parsing methods is their difficulty in handling data presented in graphical forms, such as pie charts or bar graphs. These visual elements are often used to convey complex data, but traditional parsing techniques that focus on text extraction struggle with purely graphical content. This limitation poses a significant challenge, as crucial information within these visual elements can be missed or misinterpreted. Since CarbonPDF primarily relies on text data, it cannot effectively answer questions based on content that combines graphs and text. However, our technique remains useful when numerical data is presented alongside these graphs, as it can still extract and analyze this information. In the future, we plan to explore multimodal large language models (LLMs) to perform reasoning on both text and visual data.

Although CarbonPDF can handle various types of questions, there are still limitations. For example, if CarbonPDF is asked about the carbon footprint of processors, but the exact term "processor" does not appear in the text, 
the system might not provide the expected response, 
even if related terms like "mainboard" are present. 
This occurs because the model is not capable of understanding synonyms or recognizing that certain components are subsets of larger systems. A key question for future research is whether large language models (LLMs) can be trained to handle such nuances, improving their reasoning ability to understand related terms and components within a broader context.

\section{Conclusion}
We introduce CarbonPDF-QA, an open-source product carbon footprint question-answering dataset with comprehensive annotations, comprising approximately 18,000 questions of various types derived from 1,735 product carbon reports. To facilitate extraction and reasoning, we designed CarbonPDF, a question answering model that generates executable programs to accurately answer questions directly from raw text. Our approach employs a retrieve-and-generate (RAG) pipeline, augmented by a critic model that selects the most relevant documents to improve answer accuracy. We show that CarbonPDF is able to perform accurate reasoning despite inconsistencies in the text. Through extensive experiments, we demonstrate that CarbonPDF outperforms state-of-the-art large language models and QA systems. We anticipate that both the CarbonPDF-QA dataset and the CarbonPDF model will serve as valuable benchmarks and baselines, fostering further development of advanced question-answering models tailored for PDF documents and carbon footprint estimation.

\bibliographystyle{ACM-Reference-Format}
\bibliography{paper}

\appendix

\end{document}